\documentclass{article} 
\usepackage{iclr2024_conference,times}

\usepackage{amsmath,amsfonts,bm}

\def\eqref#1{equation~\ref{#1}}

\def\1{\bm{1}}

\DeclareMathAlphabet{\mathsfit}{\encodingdefault}{\sfdefault}{m}{sl}
\SetMathAlphabet{\mathsfit}{bold}{\encodingdefault}{\sfdefault}{bx}{n}

\usepackage{url}

\iclrfinalcopy
\usepackage{latexsym}

\usepackage[T1]{fontenc}
\usepackage{tgheros}

\usepackage[utf8]{inputenc}
\usepackage{hhline}
\usepackage{multirow}

\usepackage{algorithm}
\usepackage{algpseudocode}

\usepackage{microtype}

\usepackage{inconsolata}
\usepackage{booktabs}
\usepackage{graphicx}
\usepackage{xcolor, soul}  
\usepackage{mdframed}
\usepackage{wrapfig}
\usepackage{listings}
\usepackage{savesym}
\savesymbol{checkmark}
\usepackage{dingbat}
\usepackage{tabularx}
\usepackage{amsmath,amssymb}
\definecolor{dkgreen}{rgb}{0,0.6,0}
\definecolor{gray}{rgb}{0.5,0.5,0.5}
\definecolor{mauve}{rgb}{0.58,0,0.82}
\usepackage{booktabs} 

\definecolor{cvprblue}{rgb}{0.21,0.49,0.74}
\usepackage[pagebackref,breaklinks,colorlinks,citecolor=cvprblue]{hyperref}
\definecolor{lightgray}{gray}{0.5}
\definecolor{mediumgray}{gray}{0.3}
\newcommand{\helper}{{{HELPER}}}
\newcommand{\model}{{{HELPER-X}}}
\newcommand{\modelshared}{{{HELPER-X$_{S}$}}}
\newcommand{\modelprompt}{{{HELPER-X$_{P}$}}}
\newcommand{\executor}{Executor}
\newcommand{\locator}{Locator}

\usepackage{caption}
\title{HELPER-X: A Unified Instructable Embodied Agent to Tackle Four Interactive Vision-Language Domains with Memory-Augmented Language Models}

\begin{document}
\maketitle
\vspace{-4em}
\large\noindent\makebox[1.0\textwidth][c]{
\begin{minipage}{0.25\textwidth}
\begin{center}
  Gabriel Sarch\\
  \vspace{0.5em}
\end{center}
\end{minipage}
\begin{minipage}{0.25\textwidth}
\begin{center}
  Sahil Somani$^*$\\
  \vspace{0.5em}
\end{center}
\end{minipage}
\begin{minipage}{0.25\textwidth}
\begin{center}
  Raghav Kapoor$^*$\\
  \vspace{0.5em}
\end{center}
\end{minipage}}\newline
\noindent\makebox[1.0\textwidth][c]{
\begin{minipage}{0.3\textwidth}
\begin{center}
  Michael J. Tarr \\
\end{center}
\end{minipage}
\begin{minipage}{0.3\textwidth}
\begin{center}
  Katerina Fragkiadaki\\
\end{center}
\end{minipage}}
\begin{center}
$^*$equal contribution
\end{center}
\begin{center}
Carnegie Mellon University\\
\vspace{1em}
\href{https://helper-agent-llm.github.io/}{helper-agent-llm.github.io}
\end{center}

\vspace{5mm}

\begin{abstract}
Recent research on instructable agents has used memory-augmented Large Language Models (LLMs) as task planners, a technique that retrieves language-program examples relevant to the input instruction and uses them as in-context examples in the LLM prompt to improve the performance of the LLM in inferring the correct action and task plans. In this technical report, we  extend the capabilities of HELPER, by expanding its memory with a wider array of examples and prompts, and by integrating additional APIs for asking questions.
This simple expansion of HELPER into a shared memory enables the agent to work across the domains of executing plans from dialogue, natural language instruction following, active question asking, and commonsense room reorganization. We evaluate the agent on four diverse interactive visual-language embodied agent benchmarks: ALFRED, TEACh, DialFRED, and the Tidy Task. 
\model{} achieves few-shot, state-of-the-art performance across these benchmarks using a single agent, without requiring in-domain training, and remains competitive with agents that have undergone in-domain training.
\end{abstract}

\section{Introduction}

A typical way to adapt LLMs to downstream applications is through prompting~\citep{brown2020language,alayrac2022flamingo,liu2022makes,hongjin2022selective,mishra2022cross,wei2021finetuned,song2022clip}, exploiting their strong in-context and few-shot learning abilities.
When the amount of in-context examples and task descriptions necessary to cover the task constraints increases, inference costs significantly rise due to additional attention operations. To handle computational challenges and LLM context length, a growing body of research explores the concept of ``memory-augmented prompting'' -- a method that involves retrieving a set of pertinent in-context examples to append to the prompt, thereby broadening their applicability ~\citep{perez2021true,schick2020s,gao2020making,liu2021makes,song2023llmplanner,sarch2023helper,lewis2020retrieval,mao2021generation}. \helper{}~\cite{sarch2023helper} retrieves a set of language-program examples based on the user's input instruction  and adds them to the prompt to provide  contextualized examples for GPT-4 task planning.


\begin{wrapfigure}{r}{0.25\textwidth}
    \centering
    \includegraphics[width=0.25\textwidth]{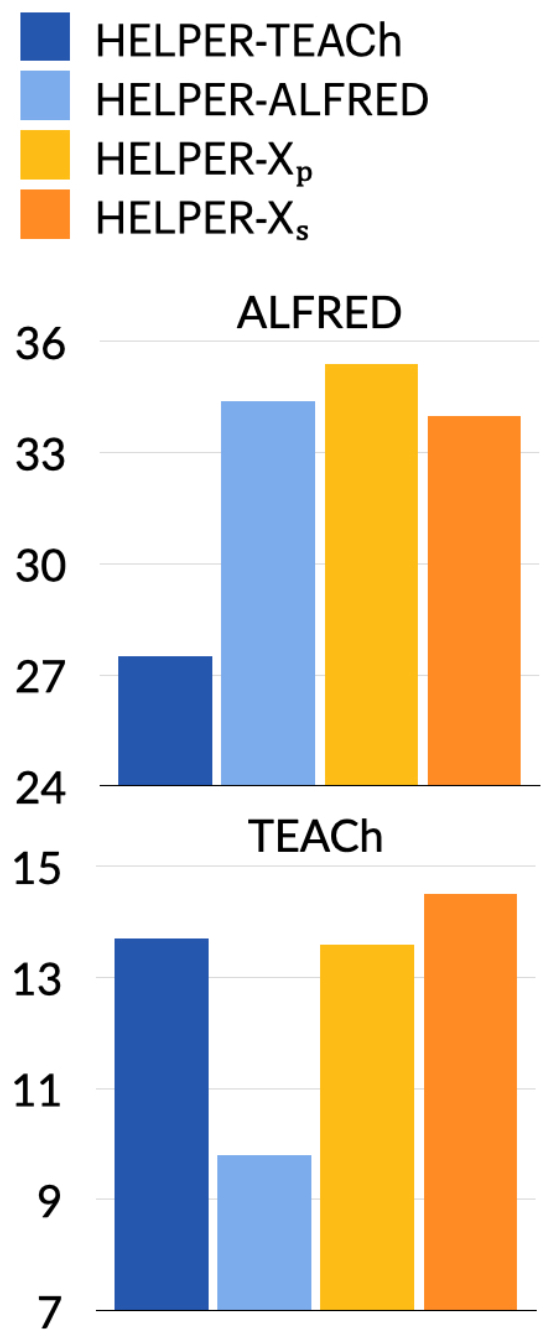}
    
    \caption{
    TEACh-tailored \helper{}~\cite{sarch2023helper} demonstrates a 6.9\% drop in success when applied to ALFRED, despite sharing the same action space and environments, due to variations in language inputs and tasks. \model{} consistently performs well in both domains with one model.
    }
    \label{fig:teaser}
    \vspace{-1 cm}
\end{wrapfigure}

Despite GPT-4's robust generalization, memory-augmented prompting tailored for one domain does not guarantee high performance in a similar yet distinct domain. 
Applying \helper{}, prompted with TEACh-specific examples and descriptions, to ALFRED -- a related domain that shares the same action space and environments but differs in language inputs and task types -- results in a notable 6.9\% decrease in accuracy compared to a \helper{} model with a specialized prompt and customized example memory specifically for ALFRED, and vice versa when doing the same on TEACh (3.2\% decrease), as shown in Figure~\ref{fig:teaser} (right).  

We report that two simple extensions of \helper{} that allows for strong performance across four domains, by expanding its memory with a wider array of examples and prompts, and by integrating a additional APIs for asking questions. Specifically, we introduce two \model{} variants: \modelprompt{}, which retrieves domain-specific prompt templates and related in-context examples for the LLM, and \modelshared{}, which retrieves in-context examples from a shared memory under a single prompt template. 


We evaluate \model{} across four domains that include dialogue-based task completion on TEACh~\citep{TEACH}, following instructions from natural language on ALFRED~\citep{shridhar2020alfred}, engaging in instruction following with active question asking on DialFRED~\citep{gao2022dialfred}, and organizing rooms using spatial commonsense reasoning in the Tidy Task~\citep{sarch2022tidee}. \model{} demonstrates state-of-the-art performance in the few-shot example domain, that is, without in-domain training. Extending the language-program memory does not cause interference and does not hinder performance. In fact, \model{} matches, and sometimes even exceeds, the performance of agents prompted with a single-domain in mind.

\section{\model{}}
We extend \helper{}~\citep{sarch2023helper} to work across four domains. We propose two versions to extend the memory-augmented prompting of LLMs in \helper{}: 1) \modelprompt{} that retrieves from a memory of domain-tailored prompt templates and associated domain-specific examples (Section~\ref{sec:prompt_ret}), and 2) \modelshared{} that expands the memory of \helper{} into a shared memory of in-context examples across domains combined with a domain-agnostic prompt template (Section~\ref{sec:shared_mem}). Additionally, we extend the capabilities of \helper{} for question asking, by appending a question API with functions defining possible questions and their arguments to the LLM prompt (Section~\ref{sec:QA}). 
We use \helper{}~\citep{sarch2023helper} for execution of the generated program using standard perception modules.

\subsection{Background}

Here, we give an account of \helper{} to make the paper self-contained. 

\helper{} prompts an LLM, namely GPT-4~\citep{gpt4technical}, to generate plans as Python programs. 
It assumes that the agent has access to a set of action skills $S$ (e.g., \texttt{go\_to(X)}, \texttt{pickup(X)}, etc.). \helper{} adds these skills to the prompt in the form of a Python API. The LLM is instructed only to call  these pre-defined skill functions in its generated programs. 
\helper{} considers a key-value memory of language inputs and successful program pairs.  It retrieves a set of  in-context examples relevant to the current input language to add to the prompt to assist  program generation.  
Each key is encoded into a language embedding. The top-$k$ language-program pairs are retrieved based on their euclidean distance to the  encoding of the language input $I$ encoding.
 The \helper{} prompt also contains a role description ("You are a helpful assistant with expertise in..."), a task description ("Your task is to ...") and guidelines to follow for program generation ("You should only use functions in the API..."), that are commonly tailored to the domain-of-interest. 

\helper{}~\citep{sarch2023helper}, using RGB input, estimates depth maps, object masks, and agent egomotion at each timestep. This facilitates the creation and upkeep of a 3D occupancy map and an object memory database, essential for obstacle navigation and object tracking. Object detection in each frame leads to instance aggregation based on 3D centroid proximity, with each instance characterized by dynamic state attributes (e.g., cooked, sliced, dirty). 
When an action fails, a Vision-Language Model (CLIP~\citep{radford2021learning}) provides feedback, prompting the LLM to re-plan. For objects not present in the map, the LLM suggests areas for \helper{} to search (e.g., ``near the sink"). 

\begin{figure*}[t!]
    \centering
    \includegraphics[width=\textwidth]{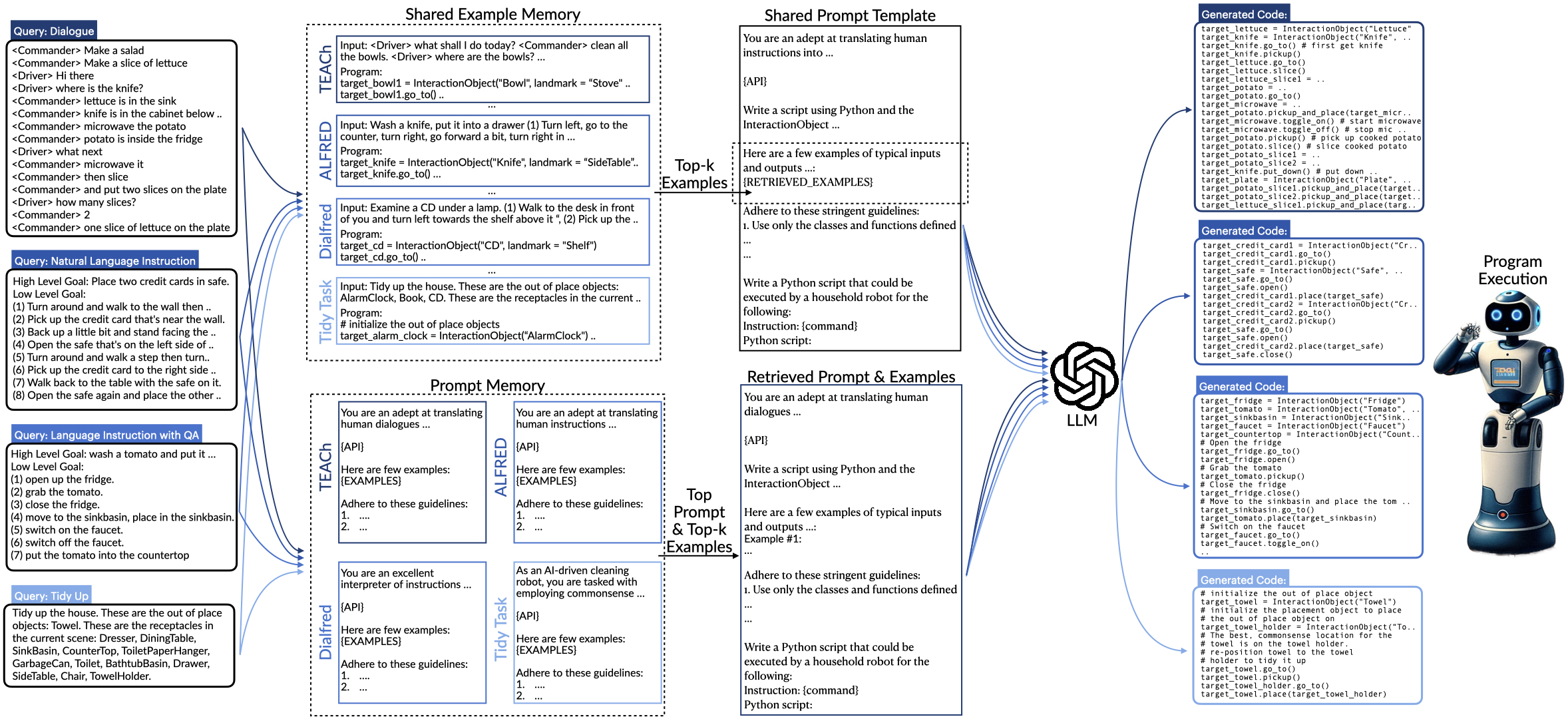}
    \caption{Illustration of the shared example memory (\modelshared{}; top) and the prompt retrieval (\modelprompt{}; bottom). The memory is shared across domains in both versions, allowing language and task inputs from any of the domains.} \label{fig:methods}
\end{figure*}

\subsection{Unified Memory-Augmented Prompting}
We explore two ways to expand \helper{} to work across four domains, either through prompt retrieval (Section~\ref{sec:prompt_ret}) or through a shared example memory (Section~\ref{sec:shared_mem}). 

\subsubsection{Prompt Retrieval} \label{sec:prompt_ret}
Given an input language instruction $I$, the prompt retrieval agent \modelprompt{} retrieves a specialized prompt template $P$ and an associated set of specialized examples $E$. Each specialized prompt template contains role descriptions, task instructions and guidelines tailored to each domain, namely,  dialogue-based task completion (based on TEACh~\citep{TEACH}), instruction following from natural language (based on ALFRED~\citep{shridhar2020alfred}), instruction following with active question asking (based on DialFRED~\citep{gao2022dialfred}), or tidying up rooms (based on Tidy Task~\citep{sarch2022tidee}). For retrieval, a query is generated from the input instruction by encoding the instruction into an embedding vector using a pre-trained language model~\citep{ada003technical}. The query retrieves the closest  key from memory, where each key represents  the language encodings of each prompt template and example text 
, as shown in Figure~\ref{fig:methods}. 
The top-$k$ in-context examples are further retrieved from the specialized set of examples associated with the retrieved prompt
and added to the retrieved prompt template, and the resulting prompt is used for LLM's program generation. 

\subsubsection{Shared Example Memory} \label{sec:shared_mem}
Given an input language instruction $I$, the shared example memory agent \modelshared{} retrieves a set of in-context examples from a shared memory  that includes in-context examples from all domains considered. These examples are added to a domain-agnostic prompt template that does not have a specialized role description, task instructions or guidelines for any single domain. A query is generated from the input instruction by encoding it into an embedding vector using a pre-trained language model~\citep{ada003technical}. The keys represent encodings of each in-context example language in the shared memory. The query embedding retrieves 
the top-$k$ nearest neighbors keys and their values. These are added to the prompt as relevant in-context examples for LLM program generation, as shown in Figure~\ref{fig:methods}.


\subsubsection{Question Asking API} \label{sec:QA}
A natural limitation of asking questions in a simulator is that only certain types of questions can be understood and answered. We constrained the set of possible questions asked by the agent by defining an API of available questions in the DialFRED~\citep{gao2022dialfred} benchmark and their arguments that \model{} can call on to gather more information. These include questions in three categories—Location, Appearance, and Direction—pertaining to the agent's next interaction object.
Importantly, this API can be continuously expanded by adding an additional function to the question-asking API. 

\noindent \textbf{Implementation details.} We follow the network implementation of \helper{}~\citep{sarch2023helper}. We use \texttt{GPT-4-0613}~\citep{gpt4technical} for text generation and \texttt{text-embedding-ada-002}~\citep{ada003technical} for text embeddings. We use the SOLQ object detector~\citep{dong2021solq} and ZoeDepth network~\citep{bhat2023zoedepth} for depth estimation from RGB input. We use $k=3$ for example retrieval.
\section{Experiments}
We test \model{} in the following benchmarks: 1. Inferring and executing action plans from dialogue  (TEACh~\citep{TEACH}), 2. Inferring and executing action plans  from instructions (ALFRED~\citep{shridhar2020alfred}), 3.  Active question asking for  seeking help during instruction execution (DialFRED~\citep{gao2022dialfred}), and 4. Tidying up rooms (Tidy Task~\citep{sarch2022tidee}). 

\subsection{Inferring and Executing Action Plans from Dialogue} \label{sec:teach}

Inferring and executing task plans from dialogue involves  understanding dialogue segments and executing related  instructions using the provided information in the dialogue. This task evaluates the agent's ability in 
 understanding noisy and free-form conversations between two humans discussing about a household task.  

\paragraph{Dataset} We use the TEACh benchmark~\citep{TEACH}, which consists of over 3,000 dialogues focused on household tasks in the AI2-THOR environment~\cite{ai2thor}. We use the Trajectory from Dialogue (TfD) variant, where an agent, given a dialogue segment, must infer action sequences to fulfill tasks like making a coffee or preparing a salad. The training dataset contains 1482 expert demonstrations with associated dialogues. The evaluation includes 181 'seen' and 612 'unseen' episodes, with 'seen' having different object placements and states than in training, and 'unseen' also having different object instances and room environments.  The agent receives an egocentric image at each step and selects actions to execute, such as \texttt{pickup(X)}, \texttt{turn\_left()}, etc. 
\vspace{1mm}

\noindent \textbf{Baselines} We consider  two kinds of baselines: \textbf{1.} Methods that supervise low-level or high-level action prediction from  language and visual input using  expert demonstrations  in the training set (1482 in number)~\citep{pashevich2021episodic,zheng2022jarvis,min2021film,zhang2022danli}, and \textbf{2.} Methods that use a small amount of expert demonstrations for prompting pretrained LLM ~\citep{sarch2023helper}. Specifically, \helper{} and  \model{}   use 11 domain-specific examples. We additionally include a comparison to \helper{}-ALF, which uses specialized prompts and examples for ALFRED. 

\noindent \textbf{Evaluation Metrics} We follow the TEACh evaluation metrics. \textbf{Task success rate (SR)} is a binary metric of whether all subtasks were successfully completed. \textbf{Goal condition success rate (GC)} quantifies the proportion of achieved goal conditions across all sessions. 
Both SR and GC have path-length weighted variants weighted by (path length of the expert trajectory) / (path length taken by the agent).

Results are reported in  Table~\ref{tab:teach_tfd}. \textbf{On validation unseen, \modelshared{} and \modelprompt{} demonstrate performance on-par with \helper{}, with \modelshared{} even slightly outperforming \helper{},} despite \model{} being shared across four domains. \model{} also outperforms the best supervised baselines trained in-domain with many demonstrations. \textbf{On validation seen,  both \model{} variants  outperform \helper{}, with the best model \modelprompt{}  outperforming \helper{} by 2.7\% in success rate.}.

\begin{table}[h!]
\begin{center}
\footnotesize
\setlength{\tabcolsep}{2pt} 
\caption{\textbf{Trajectory from Dialogue (TfD) evaluation on the TEACh validation set.}
Trajectory length weighted metrics are included in ( parentheses ). FS = few shot. Sup. = supervised. G = generalist; shared across benchmarks. GC = goal-condition success.
}\label{tab:teach_tfd}
\begin{tabular}{@{}clcccc@{}}
 \\
&  & \multicolumn{2}{c}{Unseen} & \multicolumn{2}{c}{Seen} \\
 &  & \multicolumn{1}{c}{Success} & \multicolumn{1}{c}{GC} & \multicolumn{1}{c}{Success} & \multicolumn{1}{c}{GC}
  \\ 
 \midrule
\multirow{5}{*}{\rotatebox[origin=c]{90}{\textcolor{lightgray}{Sup.}}}
 & \textcolor{lightgray}{E.T.}~\citep{pashevich2021episodic} & \textcolor{lightgray}{0.5 (0.1)} & \textcolor{lightgray}{0.4 (0.6)} & \textcolor{lightgray}{1.0 (0.2)} & \textcolor{lightgray}{1.4 (4.8)} \\
& \textcolor{lightgray}{JARVIS}~\citep{zheng2022jarvis} & \textcolor{lightgray}{1.8 (0.3)} & \textcolor{lightgray}{3.1 (1.6)} & \textcolor{lightgray}{1.7 (0.2)} & \textcolor{lightgray}{5.4 (4.5)} \\
& \textcolor{lightgray}{FILM}~\citep{min2022don} & \textcolor{lightgray}{2.9 (1.0)} & \textcolor{lightgray}{6.1 (2.5)} & \textcolor{lightgray}{5.5 (2.6)} & \textcolor{lightgray}{5.8 (11.6)} \\
& \textcolor{lightgray}{DANLI}~\citep{zhang2022danli} & \textcolor{lightgray}{8.0 (3.2)} & \textcolor{lightgray}{6.8 (6.6)} & \textcolor{lightgray}{5.0 (1.9)} & \textcolor{lightgray}{10.5 (10.3)} \\
& \textcolor{lightgray}{ECLAIR}~\citep{kim2023context} & \textcolor{lightgray}{13.2} & -- & -- & -- \\

\hline
\addlinespace[0.15cm] 
\multirow{2}{*}{\rotatebox[origin=c]{90}{\textcolor{mediumgray}{FS}}} & \textcolor{mediumgray}{\helper{}-ALF} & \textcolor{mediumgray}{10.5 (1.8)} & \textcolor{mediumgray}{10.3 (4.5)} & \textcolor{mediumgray}{13.3 (2.8)} & \textcolor{mediumgray}{14.2 (7.5)}\\
& \textcolor{mediumgray}{\helper{}}~\citep{sarch2023helper} & \textcolor{mediumgray}{13.7 (1.6)} & \textcolor{mediumgray}{\textbf{14.2 (4.6)}} & \textcolor{mediumgray}{12.2 (1.8)} & \textcolor{mediumgray}{18.6 (9.3)} \\
\addlinespace[0.15cm] 
\hline
\addlinespace[0.15cm] 
\multirow{2}{*}{\rotatebox[origin=c]{90}{G+FS}} & \modelprompt{} & 13.6 (2.0) & 13.6 (5.6) & \textbf{14.9 (3.6)} & \textbf{20.3 (11.0)} \\
& \modelshared{} & \textbf{14.5 (2.1)} & 14.0 (5.4) & 14.4 (3.5) & 19.9 (11.0) \\
\addlinespace[0.15cm] 
 \hline
\end{tabular}
\end{center}
\vspace{-17pt}
\end{table}

\subsection{Following Natural Language Instructions}~\label{sec:alfred}
Natural language instruction following evaluates the agent's ability to carry out high-level instructions (``Rinse off a mug and place it in the coffee maker") and low-level ones  (``Walk to the coffee maker on the right")  provided by a human user. Importantly, the language and tasks in this evaluation differ from the ones in the TEACh benchmark  (Section~\ref{sec:teach}). 

\paragraph{Dataset} Te ALFRED~\citep{shridhar2020alfred}  is a vision-and-language navigation benchmark designed for embodied agents to execute tasks in domestic settings from  RGB sensory input. It includes seven task types across 207 environments, that involve 115 object types in 4,703 task instances, varying from simple object relocation to  placing a heated item in a receptacle. The dataset includes detailed human-authored instructions and high-level goals, based on 21,023 expert demonstrations. It also comprises 820 'seen' and 821 'unseen' validation episodes. Agents receive egocentric RGB images at each step and select actions from a predefined set to progress, such as \texttt{pickup(X)}, \texttt{turn\_left()}, etc.

\noindent \textbf{Baselines} Again, we consider two  sets of baselines: those that supervise low-level or high-level action prediction using the expert demonstrations in the training set~\citep{pashevich2021episodic,zheng2022jarvis,min2021film,zhang2022danli,zhang2021hierarchical,song2022one,blukis2022persistent,bhambri2023multi,liu2022lebp,murray2022following,prompter2022}, and those that  use a small amount of demonstrations ($<=100$) (few-shot) ~\citep{sarch2023helper, song2023llmplanner,saycan2022arxiv,liu2023egocentric}. The SayCan~\citep{saycan2022arxiv}, FILM~\citep{min2021film} (FS), and HLSM~\citep{blukis2022persistent} (FS) few shot baselines are adapted for the few-shot ALFRED setting by the authors of \citet{song2023llmplanner}, where the planning modules in FILM and HLSM are re-trained using only 100 demonstrations. We adapt HELPER~\citep{sarch2023helper} as a baseline with specialized prompts and examples for ALFRED, as well as a comparison to \helper{}-TEACh, which uses specialized prompts and examples for TEACh. \helper{} and our \model{} model each use 7 domain-specific examples in their memory. 

\noindent \textbf{Evaluation Metrics} We follow the ALFRED evaluation metrics: \textbf{1. Task success rate (SR)} and \textbf{2. Goal condition success rate (GC)}. These are defined the same as in TEACh (Section~\ref{sec:teach}).

Results are reported in Table~\ref{tab:alfred}. 
Our conclusions are similar to Section \ref{sec:teach}. 
On validation unseen, \modelshared{} and \modelprompt{} demonstrate performance on-par with \helper{}, with \modelprompt{} marginally outperforming \helper{} by 1.0\%, despite. \model{} is also competitive with the best supervised baselines, despite only requiring a few in-domain demonstrations. On validation seen, we observe both \model{} models marginally outperforming \helper{}. We additionally show that using \helper{}-TEACh which has prompts and examples for a different domain (TEACh) causes a significant 6.9\% drop in performance.

\begin{table}[h!]
\begin{center}
\footnotesize
\setlength{\tabcolsep}{2pt} 
\caption{\textbf{Evaluation on the ALFRED validation unseen set.}
Trajectory length weighted metrics are included in ( parentheses ). FS = few shot. Sup. = supervised. G = generalist; shared across benchmarks.
}\label{tab:alfred}
\begin{tabular}{@{}llcccc@{}}
 \\
 &  & \multicolumn{2}{c}{Unseen} & \multicolumn{2}{c}{Seen} \\
 &  & \multicolumn{1}{c}{Success} & \multicolumn{1}{c}{GC} & \multicolumn{1}{c}{Success} & \multicolumn{1}{c}{GC}
  \\ 
 \midrule
\multirow{10}{*}{\rotatebox[origin=c]{90}{\textcolor{lightgray}{Sup.}}}
 & \textcolor{lightgray}{E.T.~\citep{pashevich2021episodic}} & \textcolor{lightgray}{7.3 (3.3)} & \textcolor{lightgray}{20.9 (11.3)} & \textcolor{lightgray}{46.6 (32.3)} & \textcolor{lightgray}{52.9 (42.2)} \\
& \textcolor{lightgray}{HiTUT~\citep{zhang2021hierarchical}} & \textcolor{lightgray}{12.4 (6.9)} & \textcolor{lightgray}{23.7 (12.0)} & \textcolor{lightgray}{25.2 (12.2)} & \textcolor{lightgray}{34.9 (18.5)} \\
& \textcolor{lightgray}{M-TRACK\citep{song2022one}} & \textcolor{lightgray}{17.29} & \textcolor{lightgray}{28.98} & \textcolor{lightgray}{26.70} & \textcolor{lightgray}{33.21} \\
& \textcolor{lightgray}{HLSM~\citep{blukis2022persistent}} & \textcolor{lightgray}{18.3} & \textcolor{lightgray}{31.2} & \textcolor{lightgray}{29.6} & \textcolor{lightgray}{38.7} \\
& \textcolor{lightgray}{FILM~\citep{min2021film}} & \textcolor{lightgray}{20.1} & \textcolor{lightgray}{32.5} & \textcolor{lightgray}{24.6} & \textcolor{lightgray}{37.2} \\
& \textcolor{lightgray}{MCR-Agent~\citep{bhambri2023multi}} & \textcolor{lightgray}{20.1 (10.8)} & \textcolor{lightgray}{--} & \textcolor{lightgray}{34.4 (23.0)} & \textcolor{lightgray}{--} \\
& \textcolor{lightgray}{LEBP~\citep{liu2022lebp}} & \textcolor{lightgray}{22.36} & \textcolor{lightgray}{29.58} & \textcolor{lightgray}{27.63} & \textcolor{lightgray}{35.76} \\
& \textcolor{lightgray}{LGS-RPA~\citep{murray2022following}} & \textcolor{lightgray}{33.18} & \textcolor{lightgray}{44.68} &  \textcolor{lightgray}{43.86} & \textcolor{lightgray}{52.51} \\
& \textcolor{lightgray}{EPA~\citep{liu2023egocentric}} & \textcolor{lightgray}{40.11} & \textcolor{lightgray}{44.14} & \textcolor{lightgray}{45.78} & \textcolor{lightgray}{51.03} \\
& \textcolor{lightgray}{Prompter~\citep{prompter2022}} & \textcolor{lightgray}{53.3 (19.6)} & \textcolor{lightgray}{63.0 (21.7)} & \textcolor{lightgray}{--} & \textcolor{lightgray}{--} \\
\hline
\addlinespace[0.15cm] 
\multirow{6}{*}{\rotatebox[origin=c]{90}{\textcolor{mediumgray}{FS}}} & \textcolor{mediumgray}{HLSM~\citep{blukis2022persistent} (FS)} & \textcolor{mediumgray}{0.00} & \textcolor{mediumgray}{1.86} & \textcolor{mediumgray}{0.1} & \textcolor{mediumgray}{2.8} \\
& \textcolor{mediumgray}{FILM~\citep{min2021film} (FS)} & \textcolor{mediumgray}{0.00} & \textcolor{mediumgray}{9.65} & \textcolor{mediumgray}{0.0} & \textcolor{mediumgray}{0.0} \\
& \textcolor{mediumgray}{SayCan~\citep{saycan2022arxiv}} & \textcolor{mediumgray}{9.9} & \textcolor{mediumgray}{22.5} & \textcolor{mediumgray}{12.3} & \textcolor{mediumgray}{24.5} \\
& \textcolor{mediumgray}{LLM-Planner~\citep{song2023llmplanner}} & \textcolor{mediumgray}{15.4} & \textcolor{mediumgray}{23.4} & \textcolor{mediumgray}{16.5} & \textcolor{mediumgray}{30.1} \\
& \textcolor{mediumgray}{\helper{}-TEACh} & \textcolor{mediumgray}{27.5 (5.9)} & \textcolor{mediumgray}{44.3 (10.1)} & \textcolor{mediumgray}{24.5 (6.3)} & \textcolor{mediumgray}{38.2 (11.2)} \\
& \textcolor{mediumgray}{\helper{}~\citep{sarch2023helper}} & \textcolor{mediumgray}{34.4 (7.6)} & \textcolor{mediumgray}{51.5 (11.9)} & \textcolor{mediumgray}{27.6 (7.4)} & \textcolor{mediumgray}{42.0 (12.7)} \\
\addlinespace[0.15cm] 
\hline
\addlinespace[0.15cm] 
\multirow{2}{*}{\rotatebox[origin=c]{90}{G+FS}} & \modelprompt{} & \textbf{35.4 (7.9)} & \textbf{52.9 (12.3)} & \textbf{28.2 (7.5)} & \textbf{42.5 (12.9)} \\
& \modelshared{} & 34.0 (7.5) & 51.1 (11.9) & 28.0 (7.4) & 42.1 (12.8) \\
\addlinespace[0.15cm] 
 \hline
\end{tabular}
\end{center}
\vspace{-10pt}
\end{table}

\subsection{Instruction Following with  Asking Questions}
Question asking instruction following allows the agent to choose to ask questions to an oracle to gain additional information to help it complete a task defined by an initial natural language instruction.


\paragraph{Dataset} The DialFRED benchmark~\citep{gao2022dialfred} enables an agent to query users while executing language instructions, utilizing user responses for task improvement. It features a human-annotated dataset with 53K relevant questions and answers, plus an oracle for responding to agent queries. Agents can ask questions in three categories—Location, Appearance, and Direction—pertaining to their next interaction object. The dataset covers 25 task types across 207 environments, 115 object types, and includes 'seen' and 'unseen' episodes. The agent receives egocentric RGB images at each step and selects actions from a set, like \texttt{pickup(X)}, \texttt{turn\_left()}, etc. This benchmark's instructions and tasks are distinct from TEACh and partially overlap with ALFRED, with significant modifications and 18 new task types.

\noindent \textbf{Questioning Implementation}
To ask questions in the DialFRED task, we add to the question asking API functions to query the oracle in DialFRED (see Section~\ref{sec:QA}). 
\model{} asks questions when it does not know the location of an object required for the task at hand. Unlike ALFRED, success in DialFRED requires interacting with a specific instance of an object class. To account for this, \model{} also asks questions to help disambiguate when it has seen multiple instances of the same object.  

\noindent \textbf{Baselines} We compare with the baselines in the DialFRED paper~\citep{gao2022dialfred}, which includes a sequence-to-sequence architecture for choosing to ask a question, trained with reinforcement learning, and the Episodic Transformer architecture~\citep{pashevich2021episodic}, trained with behavioral cloning. We adapt HELPER~\citep{sarch2023helper} as a baseline with specialized prompts and examples for DialFRED, as well as our question asking API. We consider a few shot setting with each few shot model receiving 7 domain-specific examples.

\noindent \textbf{Evaluation Metrics} We follow the conventions of the DialFRED benchmark. We use the \textbf{Task success rate (SR)} metric. This is defined the same as TEACh (Section~\ref{sec:teach}).

Results are reported in Table~\ref{tab:dialfred}. On validation unseen, we observe \modelshared{}  marginally outperforming \helper{} by 0.38 points in success rate, despite \model{} being shared across all domains. While \model{} is outperformed by the best supervised baselines, \model{} only requires a few in-domain demonstrations compared to the thousands of language-action demonstrations and RL interactions needed to train the baseline models. Most importantly, we see the addition of question-asking in \model{} improves success rate by 2.48 points, highlighting its efficiency in question selection and response utilization.

\begin{table}[h!]
\begin{center}

\small
\setlength{\tabcolsep}{2pt} 
\begin{minipage}{.4\linewidth}
\centering
\small
\caption{\textbf{Evaluation on the DialFRED validation unseen set.}
FS = few shot. Sup. = supervised. G = generalist; shared across benchmarks.} \label{tab:dialfred}
\begin{tabular}{@{}llcc@{}}
 \\
 &  & \multicolumn{1}{c}{Success} \\
 \midrule
\multirow{3}{*}{\rotatebox[origin=c]{90}{\textcolor{lightgray}{Sup.}}} & \textcolor{lightgray}{Instructions Only~\citep{gao2022dialfred}} & \textcolor{lightgray}{18.3} \\ 
& \textcolor{lightgray}{All QAs~\citep{gao2022dialfred}} & \textcolor{lightgray}{32.0} \\ 
& \textcolor{lightgray}{RL Anytime~\citep{gao2022dialfred}} & \textcolor{lightgray}{33.6} \\ 
\hline
\addlinespace[0.15cm] 
\multirow{1}{*}{\rotatebox[origin=c]{90}{\textcolor{mediumgray}{FS}}} & \textcolor{mediumgray}{\helper{}~\citep{sarch2023helper}} & \textcolor{mediumgray}{19.62} &\\ 
\addlinespace[0.15cm] 
\hline
\addlinespace[0.15cm] 
\multirow{3}{*}{\rotatebox[origin=c]{90}{G+FS}} & \modelprompt{} & 18.96 \\ 
& \hspace{0.5mm} \textcolor{mediumgray}{without QA} & \textcolor{mediumgray}{16.48}  \\
& \modelshared{} & \textbf{19.99} \\ 
\hline
\end{tabular}
\end{minipage}%
\hfill
\begin{minipage}{.55\linewidth}
\centering
\small
\caption{\textbf{Evaluation on the Tidy Task test set.}
Trajectory length weighted metrics are included in ( parentheses ). FS = few shot. S = supervised. G = generalist; shared across benchmarks. CM = Correctly Moved. IM = Incorrectly Moved.
}\label{tab:tidy}
\begin{tabular}{@{}clcccc@{}}
 \\
 &  & \multicolumn{1}{c}{CM $\uparrow$} & \multicolumn{1}{c}{IM $\downarrow$} & \multicolumn{1}{c}{Energy\% $\downarrow$} & {Steps}
  \\ 
 \midrule
 \addlinespace[0.15cm] 
\multirow{1}{*}{\rotatebox[origin=c]{90}{\textcolor{lightgray}{S}}}
 & \textcolor{lightgray}{TIDEE~\citep{sarch2022tidee}} & \textcolor{lightgray}{2.7} & \textcolor{lightgray}{0.3} & \textcolor{lightgray}{64.9} & 437.6\\
\addlinespace[0.15cm] 
\hline
\addlinespace[0.15cm] 
\multirow{2}{*}{\rotatebox[origin=c]{90}{\textcolor{mediumgray}{FS}}}  & \textcolor{mediumgray}{Random Receptacle} & \textcolor{mediumgray}{2.0} & \textcolor{mediumgray}{0.3} & \textcolor{mediumgray}{95.5} & 329.3\\
& \textcolor{mediumgray}{\helper{}~\citep{sarch2023helper}} & \textcolor{mediumgray}{2.1} & \textbf{\textcolor{mediumgray}{0.2}} & \textcolor{mediumgray}{83.9} & 348.3 \\

\addlinespace[0.15cm] 
\hline
\addlinespace[0.15cm] 
\multirow{2}{*}{\rotatebox[origin=c]{90}{G+FS}} & \modelprompt{} & 2.1 & 0.3 & 86.9 & 368.2\\
& \modelshared{} & \textbf{2.2} & \textbf{0.2} & \textbf{83.4} & 333.9\\
\addlinespace[0.15cm] 
 \hline
\end{tabular}
\end{minipage}
\end{center}
\end{table}

\subsection{Tidying Up using Spatial Commonsense Reasoning}
Tidying up involves figuring out where to place items without explicit instructions, relying on spatial commonsense to infer a proper location for an object. This task tests the agent's ability to use commonsense reasoning regarding contextual, object-object, and object-room spatial relations. 

\paragraph{Dataset} We  evaluate on the Tidy Task~\citep{sarch2022tidee} benchmark, where the agent is spawned in a disorganized room, and must reposition objects to bring them to an organized tidy state. 
The dataset consists of 8000 training, 200 validation, and 100 testing messy configurations in 120 distinct scenes of bedrooms, living rooms, kitchens and bathrooms.  At each time step, the agent obtains an egocentric RGB and depth image and must choose an action from a specified set to transition to the next step, such as \texttt{pickup(X)}, \texttt{turn\_left()}, etc. In this setup, the models are prompted to tidy up the room, given a set of objects that are out of place obtained using the visual detector from TIDEE~\citep{sarch2022tidee}. 


\noindent \textbf{Baselines} We compare against TIDEE~\citep{sarch2022tidee}, which includes a graph neural network encoding common object arrangements. This is supervised in the training set of the Tidy Task to predict where a target object should be re-positioned to in the current scene. We adapt HELPER~\citep{sarch2023helper} as a baseline with specialized prompts and examples for the Tidy Task. We additionally include a random receptacle baseline which chooses random receptacle placement locations for the out of place objects. We consider a few shot setting with each few shot model receiving 3 domain-specific examples.


\noindent \textbf{Evaluation Metrics}
We use the following evaluation metrics for the Tidy Task: \textbf{Correctly Moved (CM)} Average number of correctly moved objects that are out of place in the scene, and moved by the agent. Higher is better. 
\textbf{Incorrectly Moved (IM)} Average number of incorrectly moved objects that are not out of place, but were moved by the agent. Lower is better. 
\textbf{Energy} The "cleaniness" energy, where lower energy represents a higher likelihood of the room object configuration aligning with the configurations in the organized AI2THOR rooms. Following ProcThor~\citep{deitke2022️}, for each receptacle object, the probability that each object type appears on its surface is computed across the AI2THOR scenes. See the Appendix for more details. 


\vspace{2mm}

Results are in Table~\ref{tab:tidy}. On the Tidy Task, \modelshared{} and \modelprompt{} demonstrates performance on-par with \helper{}. \model{} does significantly better than if object locations are randomly placed (Random Receptacle). We find that the supervised baseline, TIDEE, outperforms \model{}, especially in the Energy metric, revealing that in-domain training on this benchmark is helpful for learning the common object configurations within the AI2THOR environments. However, we find that \model{} accomplishes the task in significantly fewer steps compared to TIDEE. 


\section{Conclusion}
We introduce \model{}, an embodied agent that executes tasks from dialogue or language instructions, ask questions, and tides up rooms. \model{} has two variants, \modelprompt{} and \modelshared{}, enhancing \helper{}'s memory capabilities. \modelprompt{} retrieves domain-specific templates and examples for large language models, while \modelshared{} retrieves only examples for a domain-agnostic prompt template through a shared memory. Evaluation of \model{} in four domains: TEACh, ALFRED, DialFRED, and the Tidy Task, yields state-of-the-art performance in the few-shot example setting. Memory and API expansions we considered maintained or improved performance for the LLM, highlighting the effectiveness of memory-enhanced LLMs in building versatile, instructable agents.

\section{Related Work}

\subsection{Memory-Augmented Prompting of Large Language Models}

Recently, external memories have been instrumental in scaling language models \cite{retro, NNLM}, overcoming the constraints of limited context windows in parametric transformers \cite{https://doi.org/10.48550/arxiv.2203.08913}. They also facilitate knowledge storage in various forms such as entity mentions \cite{DBLP:journals/corr/abs-2110-06176}, knowledge graphs \cite{https://doi.org/10.48550/arxiv.2202.10610}, and question-answer pairs \cite{https://doi.org/10.48550/arxiv.2204.04581}. Retrieval-augmented generation (RAG) \citep{lewis2020retrieval, mao2021generation} has been shown to significantly improve response quality in large language models (LLMs) by integrating external knowledge sources with the model's internal representations. In agent-based domains, memory-augmented prompting has enhanced task planning in embodied instructional contexts \citep{song2023llmplanner, sarch2023helper} and open-world gaming \citep{wang2023jarvis1, wang2023voyager, majumder2023clin}. Our model, \model{}, employs memory-augmented prompting across four benchmarks, demonstrating that memory expansion across related domains can maintain performance.

\subsection{Instructable Embodied Agents that Interact with their Environments}
Numerous benchmarks assess embodied vision-and-language tasks, with significant advancements in learning-based embodied AI agents across tasks like scene rearrangement~\cite{gan2021threedworld,RoomR,Batra2020RearrangementAC,sarch2022tidee,trabucco2022simple}, object-goal navigation~\cite{anderson2018evaluation,yang2018visual,wortsman2019learning,chaplot2020object,gupta2017cognitive,chang2020semantic,gervet2022navigating,chang2023goat}, point-goal navigation and exploration~\cite{anderson2018evaluation,savva2019habitat,wijmans2019dd,ramakrishnan2020occupancy,gupta2017cognitive,chen2019learning,chaplot2020learning,kumar2021rma}, embodied question answering~\cite{gordon2018iqa,das2018embodied,zhu2023excalibur,datta2022episodic,das2020probing,gao2022dialfred}, instructional and image navigation~\citep{ku2020room,krantz2023navigating}, audio-visual navigation~\citep{chen2020soundspaces}, interactive dialogue and natural language instruction following~\citep{yenamandra2023homerobot,shridhar2020alfred,TEACH,gao2023alexa}, and embodied commonsense reasoning~\citep{kant2022housekeep,sarch2022tidee,wu2023tidybot}. Interactive instruction benchmarks (e.g., ALFRED~\citep{shridhar2020alfred} and TEACh~\citep{TEACH}) require agents to follow natural language directives and dialogue, identifying objects in scenes via interaction masks. Variants like DialFRED~\citep{gao2022dialfred} allow agent inquiries about objects and locations. Benchmarks such as TIDEE~\citep{sarch2022tidee} and HouseKeep~\citep{kant2022housekeep} test agents' ability to tidy rooms using commonsense, without explicit object placement directives. Unlike most methods confined to a single domain, our work focuses on creating a multi-domain agent adept in dialogue-based task planning,  natural language instruction following, asking questions for disambiguation of instructions, and tidying up scenes. Our method shows competitive performance across the four domains with a few task-specific demonstrations and without domain-specific weights, beyond the single image object detector. 

Interactive vision-language embodied agent methods train distinct agents for each language-defined task, using large datasets from expert demonstrations~\citep{min2021film,prompter2022,zhang2022danli,kim2023context,pashevich2021episodic}. Some approaches use these demonstrations for end-to-end network training to directly predict actions from observations~\citep{pashevich2021episodic,gao2022dialfred,zhang2021hierarchical}. Others employ modular methods, training planners to generate subgoals handled by specialized perception, manipulation, and obstacle avoidance modules~\citep{min2021film,prompter2022,blukis2022persistent,zheng2022jarvis,kim2023context,bhambri2023multi,liu2022lebp,murray2022following,liu2023egocentric}. However, these methods often over-specialize to specific datasets and tasks, limited by the training domain's language and task structure. 
In contrast, our method performs competitively across multiple benchmarks with minimal task-specific demonstrations and without needing domain-specific networks.

\clearpage
\bibliography{custom}
\bibliographystyle{iclr2024_conference}

\appendix
\makeatletter
\clearpage
\renewcommand \thesection{S\@arabic\c@section}
\renewcommand\thetable{S\@arabic\c@table}
\renewcommand \thefigure{S\@arabic\c@figure}
\renewcommand \thelstlisting{S\@arabic\c@lstlisting}
\renewcommand \thealgorithm{S\@arabic\c@algorithm}
\makeatother

\setcounter{section}{0}
\setcounter{figure}{0}  
\setcounter{table}{0} 

\renewcommand{\theHsection}{Supplement.\thesection}
\renewcommand{\theHtable}{Supplement.\thetable}
\renewcommand{\theHfigure}{Supplement.\thefigure}

\section{Limitations}
\noindent Our model has the following limitations:




\textbf{1. Task planning from multimodal input:} 
Currently, our LLM receives the environment's state only in case of a failure, through VLM feedback. 
Integrating the visual state of the environment in a more direct way may dramatically  increase the accuracy of predicted plans.  
This direction aligns with recent work~\citep{wang2023jarvis1,mu2023embodiedgpt,yang2023octopus} that uses visual features as input to language models. 

\textbf{2. Cost of GPT-4:} While GPT-4 is the most accurate Large Language Model, its high cost necessitates exploring alternatives such as open-source models, hardware optimization,  model compression or distillation of its knowledge to smaller models, to reduce expenses.

\textbf{3. Manual Addition of Domains:} Our model supports four domains with shared examples and prompts, but manual intervention is needed for adding significantly different domains and tasks. Future developments should focus on automating the detection and integration of out-of-domain inputs.

\section{Prompts} \label{sec:prompts}


\subsection{Prompt templates for prompt retrieval} \label{prompts}
In the prompt retrieval experiments, we include four prompt templates to be retrieved. These templates are shown for TEACh, ALFRED, Dialfred, and the Tidy Task in Listing~\ref{prompt_template_teach}, Listing~\ref{prompt_template_alfred}, Listing~\ref{prompt_template_dialfred}, Listing~\ref{prompt_template_tidytask}, respectively. 

\subsection{In-Context Examples} \label{examples}
Samples of the in-context examples are shown for TEACh, ALFRED, Dialfred, and the Tidy Task in Listing~\ref{example_teach}, Listing~\ref{example_alfred}, Listing~\ref{example_dialfred}, Listing~\ref{example_tidy_task}, respectively. 

\section{Question Asking} \label{QA_supp}
\subsection{Overview}
In the DialFRED benchmark, when \model{} is unable to find an object, it is able to ask one of three question types in order to aid itself. In a real-world scenario, \model{} could take advantage of the LLM's capability to ask many types of questions, but the DialFRED benchmark limits us to three: direction, location, and appearance. 

\subsection{Question asking pipeline}
When \model{} does not have an object's location already in its memory or multiple instances of an objects exist in the memory, it forms a prompt with its current context and the API of available questions, as in Listing~\ref{prompt_template_question}. Based on the context, \model{} then chooses and asks the most appropriate question. The returned answer and an API of search related actions, alongside the context and question, are then formed into another prompt, seen in Listing~\ref{prompt_template_answer}. Finally, this prompt is parsed by \model{} into an action script to search for the object. Examples of this full pipeline for if an object does not exist in the memory are in Listing~\ref{example_QA_1} and Listing~\ref{example_QA_2}.

\section{Pre-conditions} \label{sec:precond_app}
An example of a pre-condition check for a macro-action is provided in Listing~\ref{precond_example}.

\section{Example LLM inputs \& Outputs} \label{sec:exampleLLM}
We provide examples of dialogue input, retrieved examples, and LLM output for a TEACh sample in Listing~\ref{example1}, Listing~\ref{example2}, and Listing~\ref{example3}.

\section{Simulation environment} \label{Simulator}
The TEACh dataset builds on the Ai2thor simulation environment~\citep{ai2thor}. At each time step the agent may choose from the following actions: Forward(), Backward(), Turn Left(), Turn Right(), Look Up(), Look Down(), Strafe Left(), Strafe Right(), Pickup(X), Place(X), Open(X), Close(X), ToggleOn(X), ToggleOff(X), Slice(X), and Pour(X), where X refers an object specified via a relative coordinate $(x, y)$ on the egocentric RGB frame. Navigation actions move the agent in discrete steps. We rotate in the yaw direction by 90 degrees, and rotate in the pitch direction by 30 degrees. The RGB and depth sensors are at a resolution of 480x480, a field of view of 90 degrees, and lie at a height of 0.9015 meters. The agent's coordinates are parameterized by a single $(x,y,z)$ coordinate triplet with $x$ and $z$ corresponding to movement in the horizontal plane and $y$ reserved for the vertical direction. The TEACh benchmark allows a maximum of 1000 steps and 30 API failures per episode.

\section{$\executor$ details}

\subsection{Semantic mapping and planning}

\paragraph{Obstacle map} 
\model{} maintains a 2D overhead occupancy map of its environment $\in \mathbb{R}^{H \times W}$ that it updates at each time step from the input RGB-D stream. The map is used for exploration and navigation in the environment. 

At every time step $t$, we unproject the input depth maps using intrinsic and extrinsic information of the camera to obtain a 3D occupancy map registered to the coordinate frame of the agent, similar to earlier navigation agents \cite{chaplot2020learning}. The 2D overhead maps of obstacles and free space are computed by projecting the 3D occupancy along the height direction at multiple height levels and summing.
For each input RGB image, we run a SOLQ object segmentor~\citep{dong2021solq} (pretrained on COCO~\cite{lin2014microsoft} then finetuned on TEACh rooms) to localize each of 116 semantic object categories. 
For failure detection, we use a simple matching approach from \citet{min2021film} to compare RGB pixel values before and after taking an action. 

\paragraph{Object location and state tracking}

We maintain an object memory as a list of object detection 3D centroids and their predicted semantic labels $\{ [ (X,Y,Z)_i, \ell_i\in\{1...N\} ] , i=1..K  \}  $, where $K$ is the number of objects detected thus far.
The object centroids are expressed with respect to the coordinate system of the agent, and, similar to the semantic maps, updated over time using egomotion. We track previously detected objects by their 3D centroid $C \in \mathbb{R}^{3}$. We estimate the centroid by taking the 3D point corresponding to the median depth within the segmentation mask and bring it to a common coordinate frame. We do a simple form of non-maximum suppression on the object memory, by comparing the euclidean distance of centroids in the memory to new detected centroids of the same category, and keep the one with the highest score if they fall within a distance threshold.

For each object in the object memory, we maintain an object state dictionary with a pre-defined list of attributes. These attributes include: category label, centroid location, holding, detection score, can use, sliced, toasted, clean, cooked. For the binary attributes, these are initialized by sending the object crop, defined by the detector mask, to the VLM model, and checking its match to each of [f"The \{object\_category\} is \{attribute\}", f"The \{object\_category\} is not \{attribute\}"]. We found that initializing these attributes with the VLM gave only a marginal difference to initializing them to default values in the TEACh benchmark, so we do not use it for the TEACh evaluations. However, we anticipate a general method beyond dataset biases of TEACh would much benefit from such vision-based attribute classification.

\begin{table*}[h!] 
\begin{center}
\scriptsize
\setlength{\tabcolsep}{2pt} 
\caption{\textbf{Alternative TEACh Execution from Dialog History
(EDH) evaluation split.}
Trajectory length weighted metrics are included in ( parentheses ). SR = success rate. GC = goal condition success rate. Note that Test Seen and Unseen are not the true TEACh test sets, but an alternative split of the validation set used until the true test evaluation is released, as mentioned in the TEACh github README, and also reported by DANLI~\citep{zhang2022danli}.
}\label{app:edh_alt}
\begin{tabular}{@{}llllllllll@{}}
 & \multicolumn{4}{c}{\textbf{Validation}} & \multicolumn{4}{c}{\textbf{Test}} \\
 \cmidrule(lr){2-5} \cmidrule(lr){6-9}
 & \multicolumn{2}{c}{\textbf{Unseen}} & \multicolumn{2}{c}{\textbf{Seen}} & \multicolumn{2}{c}{\textbf{Unseen}} & \multicolumn{2}{c}{\textbf{Seen}} \\
  & \multicolumn{1}{c}{SR} & \multicolumn{1}{c}{GC} & \multicolumn{1}{c}{SR} & \multicolumn{1}{c}{GC} & \multicolumn{1}{c}{SR} & \multicolumn{1}{c}{GC} & \multicolumn{1}{c}{SR} & \multicolumn{1}{c}{GC} \\ 
 \midrule
E.T. & 8.35 (0.86) & 6.34 (3.69) & 8.28 (1.13) & 8.72 (3.82) & 7.38 (0.97) & 6.06 (3.17) & 8.82 (0.29) &  9.46 (3.03) \\
DANLI & \textbf{17.25} (7.16) & 23.88 (19.38) & 16.89 (9.12) & 25.10 (22.56) & 16.71 (7.33) & 23.00 (20.55) & \textbf{18.63} (9.41) & 24.77 (21.90) \\
\textsc{HELPER} & \textbf{17.25} (3.22) & \textbf{25.24} (8.12) & \textbf{19.21} (4.72) & \textbf{33.54} (10.95) & \textbf{17.55} (2.59) & \textbf{26.49} (7.67) & 17.97 (3.44) & \textbf{30.81} (8.93) \\
 \hline
\end{tabular}
\end{center}

\vspace{-17pt}
\end{table*}

\paragraph{Exploration and path planning} 
$\model$ explores the scene using a classical mapping method. We take the initial position of the agent to be the center coordinate in the map. We rotate the agent in-place and use the observations to instantiate an initial map. Second, the agent incrementally completes the maps by randomly sampling an unexplored, traversible location based on the 2D occupancy map built so far, and then navigates to the sampled location, accumulating the new information into the maps at each time step. The number of observations collected at each point in the 2D occupancy map is thresholded to determine whether a given map location is explored or not. 
Unexplored positions are sampled until the environment has been fully explored, meaning that the number of unexplored points is fewer than a predefined threshold.

To navigate to a goal location, we compute the geodesic distance to the goal from all map locations using graph search~\cite{inoue2022prompter} given the top-down occupancy map and the goal location in the map. We then simulate action sequences and greedily take the action sequence which results in the largest reduction in geodesic distance.

\subsection{2D-to-3D unprojection}\quad For the $i$-th view, a 2D pixel coordinate $(u,v)$ with depth $z$ is unprojected and transformed to its coordinate $(X,Y,Z)^T$ in the reference frame:
\begin{equation}
    (X,Y,Z,1) = \mathbf{G}_{i}^{-1} \left(z \frac{u-c_{x}}{f_{x}}, z \frac{v-c_{y}}{f_{y}}, z, 1\right)^{T}
\end{equation}
where $(f_x, f_y)$ and $(c_x, c_y)$ are the focal lengths and center of the pinhole camera model and $\mathbf{G}_i \in SE(3)$ is the camera pose for view $i$ relative to the reference view. This module unprojects each depth image $I_i \in \mathbb{R}^{H\times W \times3}$ into a pointcloud in the reference frame $P_i \in \mathbb{R}^{M_i \times 3}$ with $M_i$ being the number of pixels with an associated depth value. 



\section{Additional details of the Tidy Task} \label{sec:tidy_metrics_app}
\subsection{Metric Definitions in the Tidy Task} 
The metrics in the original TIDEE paper~\citep{sarch2022tidee} require separate human evaluations on Amazon Mechanical Turk. We define a new set of metrics that does not require expensive annotations from humans for every evaluation. Below are detailed descriptions of each of the new metrics:
\begin{enumerate}
\item \textbf{Correctly Moved (CM)} Average number of correctly moved objects that are out of place in the scene, and moved by the agent. Higher is better. 
\item \textbf{Incorrectly Moved (IM)} Average number of incorrectly moved objects that are not out of place, but were moved by the agent. Lower is better. 
\item \textbf{Energy} Following ProcThor~\citep{deitke2022️}, for each receptacle object, the probability that each object type appears on its surface is computed across the AI2THOR scenes. Here, we compute the total number of times each object type is on the receptacle type and divide it by the total number of times the receptacle type appears across the scenes. The energy metric in the Tidy Task is defined as follows: \\
\begin{equation}
(P_{cleanup} - P_{original}) / (P_{dirty} - P_{original}) 
\end{equation}
where $P_{cleanup}$, $P_{dirty}$, and $P_{original}$ represent the sum of the object location probabilities for the cleaned up state of the room, the dirty/messy state of the room, and the original state of the room with objects put in-place by human designers, respectively. Lower is better. 
\item \textbf{Steps} Average number of steps taken by the agent per episode. 
\end{enumerate}

\subsection{Langauge Instructions for the Tidy Task} 
Since the Tidy Task does not include natural language instruction annotations, we formulate the language instruction as the following to give to the \helper{} baseline and \model{}: \textit{“Tidy up the house. These are the out of place objects: \{detected\_out\_of\_place\_objects\}. These are the receptacles in the current scene: \{detected\_receptacles\}”}, where \{detected\_out\_of\_place\_objects\} are the objects classified as out of place, and \{detected\_receptacles\} are any receptacle detected in the scene by the agent. 

To obtain the list of out of place objects, we allow the agents use of the TIDEE~\citep{sarch2022tidee} visual detector to determine whether each object detected during the mapping phase is out of place. We found that out of place detection benefits significantly from visual detection in the Tidy Task, and thus we do not use an LLM for detecting the out of place attribute. Notably, adding the additional out of place attribute to the objects in the object memory can be shared across all benchmarks.

\lstset{escapeinside={<@}{@>}, language=}
\onecolumn\begin{lstlisting}[caption={Prompt template for TEACh},captionpos=t,label={prompt_template_teach}] 
You are an adept at translating human dialogues into sequences of actions for household robots. Given a dialogue between a <Driver> and a <Commander>, you convert the conversation into a Python program to be executed by a robot.

{API}

Write a script using Python and the InteractionObject class and functions defined above that could be executed by a household robot. 

Here are a few examples of typical inputs and outputs (only for in-context reference):
{RETRIEVED_EXAMPLES}

Adhere to these stringent guidelines:
1. Use only the classes and functions defined previously. Do not create functions that are not provided above.
2. Make sure that you output a consistent plan. For example, opening of the same object should not occur in successive steps.
3. Make sure the output is consistent with the proper affordances of objects. For example, a couch cannot be opened, so your output should never include the open() function for this object, but a fridge can be opened. 
4. The input is dialogue between <Driver> and <Commander>. Interpret the dialogue into robot actions. Do not output any dialogue.
5. Object categories should only be chosen from the following classes: {OBJECT_CLASSES}
6. You can only pick up one object at a time. If the agent is holding an object, the agent should place or put down the object before attempting to pick up a second object.
7. Each object instance should instantiate a different InteractionObject class even if two object instances are the same object category. 
Follow the output format provided earlier. Think step by step to carry out the instruction.

Write a Python script that could be executed by a household robot for the following:
dialogue: {command}
Python script: 
\end{lstlisting}

\lstset{escapeinside={<@}{@>}, language=}
\onecolumn\begin{lstlisting}[caption={Prompt template for ALFRED},captionpos=t,label={prompt_template_alfred}] 
You are an excellent interpreter of instructions for household tasks. Given a task overview <High Level Goal> and step to perform <Low Level Goal>, you break the instructions down into a sequence of robotic actions.

{API}

Write a script using Python and the InteractionObject class and functions defined above that could be executed by a household robot. 

Here are a few examples of typical inputs and outputs (only for in-context reference):
{RETRIEVED_EXAMPLES}

Adhere to these stringent guidelines:
1. Use only the classes and functions defined previously. Do not create functions that are not provided above.
2. Make sure that you output a consistent plan. For example, opening of the same object should not occur in successive steps.
3. Make sure the output is consistent with the proper affordances of objects. For example, a couch cannot be opened, so your output should never include the open() function for this object, but a fridge can be opened. 
4. The input is high level task description and low level subgoals to perform the high level task. Interpret the instructions into robot actions.
5. Object categories should only be chosen from the following classes: {OBJECT_CLASSES}
6. You can only pick up one object at a time. If the agent is holding an object, the agent should place or put down the object before attempting to pick up a second object.
7. Each object instance should instantiate a different InteractionObject class even if two object instances are the same object category. 
8. Always focus on solving the high level goal. Low level instructions should only be used to guide and plan better.
9. Before performing each action, check if that action is allowed for a particular receptacle class. A few examples have been given in API documentation.
10. Check if the receptacle needs to be opened before placing the object. If yes, then open the receptacle before placing the object.
Follow the output format provided earlier. Think step by step to carry out the instruction.

Write a Python script that could be executed by a household robot for the following:
{command}
Python script: 
\end{lstlisting}

\lstset{escapeinside={<@}{@>}, language=}
\onecolumn\begin{lstlisting}[caption={Prompt template for the Dialfred},captionpos=t,label={prompt_template_dialfred}] 
You are an excellent interpreter of instructions for household tasks. Given a task overview <High Level Goal> and step to perform <Low Level Goal>, you break the instructions down into a sequence of robotic actions.

{API}

Write a script using Python and the InteractionObject class and functions defined above that could be executed by a household robot. 

Here are a few examples of typical inputs and outputs (only for in-context reference):
{RETRIEVED_EXAMPLES}

Adhere to these stringent guidelines:
1. Use only the classes and functions defined previously. Do not create functions that are not provided above.
2. Make sure that you output a consistent plan. For example, opening of the same object should not occur in successive steps.
3. Make sure the output is consistent with the proper affordances of objects. For example, a couch cannot be opened, so your output should never include the open() function for this object, but a fridge can be opened. 
4. The input is high level task description and low level subgoals to perform the high level task. Interpret the instructions into robot actions.
5. Object categories should only be chosen from the following classes: {OBJECT_CLASSES}
6. You can only pick up one object at a time. If the agent is holding an object, the agent should place or put down the object before attempting to pick up a second object.
7. Each object instance should instantiate a different InteractionObject class even if two object instances are the same object category. 
8. Make sure that you are solving both the high level goal and the low level goals. Some instructions may only be present in one or the other, so address everything from both.
9. Before performing each action, check if that action is allowed for a particular receptacle class. A few examples have been given in API documentation.
Follow the output format provided earlier. Think step by step to carry out the instruction.

Write a Python script that could be executed by a household robot for the following:
{command}
Python script: 
\end{lstlisting}

\lstset{escapeinside={<@}{@>}, language=}
\onecolumn\begin{lstlisting}[caption={Prompt template for the Tidy Task},captionpos=t,label={prompt_template_tidytask}] 
Task: As an AI-driven cleaning robot, you are tasked with employing commonsense reasoning to identify where to place out of place objects that aren't situated appropriately. Given a list of out of place objects, you are to write a Python program to be executed by a robot that will bring the out of place objects to a suitable location.

{API}

Write a script using Python and the InteractionObject class and functions defined above that could be executed by a household robot. 

Here are a few examples of typical inputs and outputs (only for in-context reference):
{RETRIEVED_EXAMPLES}

Adhere to these stringent guidelines:
1. Use only the classes and functions defined previously. Do not create functions that are not provided above.
2. Make sure that you output a consistent plan. For example, opening of the same object should not occur in successive steps.
3. Make sure the output is consistent with the proper affordances of objects. For example, a couch cannot be opened, so your output should never include the open() function for this object, but a fridge can be opened. 
4. Object categories should only be chosen from the following classes: {OBJECT_CLASSES}
5. You can only pick up one object at a time. If the agent is holding an object, the agent should place or put down the object before attempting to pick up a second object.
6. Each object instance should instantiate a different InteractionObject class even if two object instances are the same object category. 
7. Address each item systematically, one by one.
8. Base your decisions on your ingrained knowledge about the typical placement of day-to-day objects.
Follow the output format provided earlier. Think step by step to carry out the instruction.

Write a Python script that could be executed by a household robot for the following:
input: {command}
Python script: 
\end{lstlisting}

\lstset{escapeinside={<@}{@>}, language=Python}
\onecolumn\begin{lstlisting}[caption={Sample in-context example for TEACh},captionpos=t,label={example_teach}] 
<@\textcolor{orange}{ 
Dialogue input:}@>

<Driver> what shall I do today? <Commander> clean all the bowls. <Driver> where are the bowls? <Commander> start with the one by the stove. <Commander> left. <Commander> rinse it with water. <Commander> great. <Driver> what next? <Commander> the next one is in the fridge. <Commander> you need to rinse it with water also. <Commander> great job. we are finished. 

<@\textcolor{red}{Python script:}@>
target_bowl1 = InteractionObject("Bowl", landmark = "Stove", attributes = ["clean"])
target_bowl1.go_to()
target_bowl1.pickup()
target_bowl1.clean()
target_bowl1.put_down()
target_bowl2 = InteractionObject("Bowl", landmark = "Fridge", attributes = ["clean"])
target_bowl2.go_to()
target_bowl2.pickup()
target_bowl2.clean()
target_bowl2.put_down()
\end{lstlisting}

\lstset{escapeinside={<@}{@>}, language=Python}
\onecolumn\begin{lstlisting}[caption={Sample in-context example for ALFRED},captionpos=t,label={example_alfred}] 
<@\textcolor{orange}{ 
High Level Goal: To heat an apple and place in the black bin. }@>
<@\textcolor{orange}{ Low Level Goal:}@>
<@\textcolor{orange}{ (1) Turn around and walk to the kitchen island. }@>
<@\textcolor{orange}{ (2) Pick up the apple in front of the gold colored plate. }@>
<@\textcolor{orange}{ (3) Walk around the kitchen island and to the stove on the right, look above the stove to }@>
<@\textcolor{orange}{face the microwave. }@>
<@\textcolor{orange}{ (4) Place the apple inside the microwave, heat up/cook the apple, take the apple out of the }@>
<@\textcolor{orange}{microwave. }@>
<@\textcolor{orange}{ (5) Turn left, turn left at the fridge, turn left to face the kitchen island, and look down}@>
<@\textcolor{orange}{at the black bin. }@>
<@\textcolor{orange}{ (6) Place the apple in the bin on the right side. }@>
<@\textcolor{red}{Python script:}@>
target_apple = InteractionObject("Apple", landmark = "CounterTop")
target_apple.go_to()
target_apple.pickup()
target_microwave = InteractionObject("Microwave")
target_microwave.go_to()
target_microwave.open() # open microwave before placing
target_apple.place(target_microwave) 
target_microwave.close() # close microwave before toggle on
target_microwave.toggle_on() # toggle on to heat up & cook
target_microwave.toggle_off() # Important! toggle off to end heating. Do not try to open microwave without toggle off!
target_microwave.open() # open microwave before picking
target_apple.pickup()
target_trashcan = InteractionObject("GarbageCan")
target_trashcan.go_to()
target_apple.place(target_trashcan)
\end{lstlisting}

\lstset{escapeinside={<@}{@>}, language=Python}
\onecolumn\begin{lstlisting}[caption={Sample in-context example for Dialfred},captionpos=t,label={example_dialfred}] 
<@\textcolor{orange}{ High Level Goal: Put the watch in the bowl on the nightstand.}@>
<@\textcolor{orange}{ Low Level Goal:}@>
<@\textcolor{orange}{ (1) Turn around, walk around the bookshelf, turn to the desk.}@>
<@\textcolor{orange}{ (2) Pick the watch up off of the desk.}@>
<@\textcolor{orange}{ (3) Put the watch in the bowl on the desk.}@>
<@\textcolor{orange}{ (4) Pick up the bowl off of the desk.}@>
<@\textcolor{orange}{ (5) Walk back around the bookshelf, walk over between the two beds.}@>
<@\textcolor{orange}{ (6) Put the bowl on the nightstand.}@>
<@\textcolor{red}{Python script:}@>
target_watch = InteractionObject("Watch", landmark = "Desk")
target_watch.go_to()
target_watch.pickup()
target_bowl = InteractionObject("Bowl", landmark = "Desk")
target_bowl.go_to()
target_watch.place(target_bowl)
target_bowl.go_to()
target_bowl.pickup()
target_nightstand = InteractionObject("SideTable", landmark = "Bed")
target_nightstand.go_to()
target_bowl.place(target_nightstand)
\end{lstlisting}

\lstset{escapeinside={<@}{@>}, language=Python}
\onecolumn\begin{lstlisting}[caption={Sample in-context example for the Tidy Task},captionpos=t,label={example_tidy_task}] 
<@\textcolor{orange}{input: Tidy up the house. These are the out of place objects: Potato, Knife. These are }@>
<@\textcolor{orange}{the receptacles in the current scene: DiningTable, Microwave, CoffeeMachine, CounterTop.}@>
<@\textcolor{red}{Python script:}@>
# initialize the out of place objects
target_potato = InteractionObject("Potato")
target_knife = InteractionObject("Knife")
# initialize the placement objects to place the out of place object on
target_countertop = InteractionObject("CounterTop") # The best, commonsense location for both the potato and knife is on the countertop.
# re-position potato to the countertop to tidy it up
target_potato.go_to()
target_potato.pickup()
target_countertop.go_to()
target_potato.place(target_countertop)
# re-position knife to the countertop to tidy it up
target_knife.go_to()
target_knife.pickup()
target_countertop.go_to()
target_knife.place(target_countertop)
\end{lstlisting}

\lstset{escapeinside={<@}{@>}, language=}
\onecolumn\begin{lstlisting}[caption={Prompt template for Question Selection},captionpos=t,label={prompt_template_question}] 
You are an excellent interpreter of human instructions for household tasks. Given a list of questions you can ask and information about the current environment and context, you provide a question that should be asked in order to give the agent useful information.

{API}

Write a script using Python using the class and functions defined above that could be executed by a household robot. 

Adhere to these stringent guidelines:
1. Use only the classes and functions defined previously. Do not create functions that are not provided above.
2. Make sure you choose the question that provides the most information and is most relevant for the situation at hand.
3. Object categories should only be chosen from the following classes: {OBJECT_CLASSES}
Follow the output format provided earlier. Think step by step to carry out the instruction.

Write a Python script that asks questions to help a household robot in the following situation:
{context}
Python script: 
\end{lstlisting}

\lstset{escapeinside={<@}{@>}, language=}
\onecolumn\begin{lstlisting}[caption={Prompt template for Answer Parsing},captionpos=t,label={prompt_template_answer}] 
You are an excellent interpreter of human instructions for household tasks. Given the current context of the agent, a question that was asked, and an answer that was given, you must write code for actions the agent should take based on the answer provided.

{API}

Write a script using Python using the class and functions defined above that could be executed by a household robot. 

Adhere to these stringent guidelines:
1. Use only the classes and functions defined previously. Do not create functions that are not provided above.
2. Make sure you plan the most simple and direct interpretation of the answer given.
3. Prioritize the most specific information given. For example, an actual object name should be deemed more important than a region.
4. If multiple pieces of information are given, ensure you incorporate all of them into the script.
5. Object categories should only be chosen from the following classes: {OBJECT_CLASSES}

Write a Python script that asks questions to help a household robot in the following situation:
{context}
{question}
{answer}
Python script: 
\end{lstlisting}

\lstset{escapeinside={<@}{@>}, language=Python}
\onecolumn\begin{lstlisting}[caption={Sample in-context example 1 for Question Asking},captionpos=t,label={example_QA_1}] 
<@\textcolor{orange}{Question Selection Input: }@>
<@\textcolor{orange}{Context: The agent does not know where the ButterKnife is.}@>
<@\textcolor{red}{Questioning Script:}@>
askForLocation('ButterKnife')

<@\textcolor{orange}{Answer Parsing Input: }@>
<@\textcolor{orange}{Context: The agent does not know where the ButterKnife is. }@>
<@\textcolor{orange}{Question Asked: askForLocation('ButterKnife')}@>
<@\textcolor{orange}{Answer Returned: The ButterKnife is to your left on the countertop.}@>

<@\textcolor{red}{Parsed Answer:}@>
# Turn to the left as per the instruction
turn('left')
# Search for the butterknife on the counter top
search_near_other_object('ButterKnife', 'CounterTop')
\end{lstlisting}

\lstset{escapeinside={<@}{@>}, language=Python}
\onecolumn\begin{lstlisting}[caption={Sample in-context example 2 for Question Asking},captionpos=t,label={example_QA_2}] 
<@\textcolor{orange}{Question Selection Input:}@>
<@\textcolor{orange}{Context: The agent does not know where the SoapBar is.}@>
<@\textcolor{red}{Questioning Script:}@>
askForLocation('SoapBar')

<@\textcolor{orange}{Answer Parsing Input:}@>
<@\textcolor{orange}{Context: The agent does not know where the SoapBar is.}@>
<@\textcolor{orange}{Question Asked: askForLocation('SoapBar')}@>
<@\textcolor{orange}{Answer Returned: The SoapBar is to your front right in the garbage can.}@>
<@\textcolor{red}{Parsed Answer:}@>
# Turn right as the SoapBar is to the front right
turn('right')
# Move forward to reach the garbage can
move('forward')
# Search for the SoapBar near the garbage can
search_near_other_object('SoapBar', 'GarbageCan')


\end{lstlisting}

\clearpage





\end{document}